\documentclass[letterpaper]{article} 
\usepackage{aaai24}  
\usepackage{times}  
\usepackage{helvet}  
\usepackage{courier}  
\usepackage[hyphens]{url}  
\usepackage{graphicx} 
\usepackage{natbib}  
\usepackage{caption} 
\usepackage{algorithm}
\usepackage{algorithmic}
\usepackage{amsfonts}
\usepackage{amsmath}
\usepackage{makecell}
\usepackage{enumitem} 
\usepackage{multirow}
\usepackage{pifont}
\newcommand{\cmark}{\ding{51}}%
\newcommand{\xmark}{\ding{55}}%

\usepackage{newfloat}
\usepackage{listings}
\DeclareCaptionStyle{ruled}{labelfont=normalfont,labelsep=colon,strut=off} 
\floatstyle{ruled}
\newfloat{listing}{tb}{lst}{}
\floatname{listing}{Listing}
\title{DVANet: Disentangling View and Action Features for Multi-View Action Recognition}
\author {
    Nyle Siddiqui,
    Praveen Tirupattur,
    Mubarak Shah
}
\affiliations {
    Center for Research in Computer Vision\\
    University of Central Florida\\
    Orlando, Florida, USA, 32816\\
    nyle.siddiqui@ucf.edu, praveen.tirupattur@ucf.edu, shah@crcv.ucf.edu
}

\begin{document}

\maketitle

\begin{abstract}
 In this work, we present a novel approach to multi-view action recognition where we guide learned action representations to be separated from view-relevant information in a video. When trying to classify action instances captured from multiple viewpoints, there is a higher degree of difficulty due to the difference in background, occlusion, and visibility of the captured action from different camera angles. To tackle the various problems introduced in multi-view action recognition, we propose a novel configuration of learnable transformer decoder queries, in conjunction with two supervised contrastive losses, to enforce the learning of action features that are robust to shifts in viewpoints. Our disentangled feature learning occurs in two stages: the transformer decoder uses separate queries to separately learn action and view information, which are then further disentangled using our two contrastive losses. We show that our model and method of training significantly outperforms all other uni-modal models on four multi-view action recognition datasets: NTU RGB+D, NTU RGB+D 120, PKU-MMD, and N-UCLA. Compared to previous RGB works, we see maximal improvements of 1.5\%, 4.8\%, 2.2\%, and 4.8\% on each dataset, respectively.
\end{abstract}

\section{Introduction}
\label{intro}

Human action recognition (HAR) has played an integral role in the development and progression of computer vision research. HAR has established itself as a critical intersection between academic research and real-world implementation due to its practical application to a myriad of domains, such as human-computer interaction \cite{chakraborty2018review}, road safety \cite{ismail2010application}, video understanding \cite{hussain2021comprehensive}, security/surveillance \cite{black2002multi}, and more. Consequently, there are a plethora of large-scale datasets that have been collected over the years to provide ample resources for researchers to advance this historically active field. Furthermore, HAR datasets have grown in difficulty proportionally to the astounding capabilities of deep learning in computer vision, such as the introduction of additional modalities, increased number of subjects and actions, and multiple viewpoints. 

\begin{figure*}
    \centering
    \includegraphics[width=\linewidth]{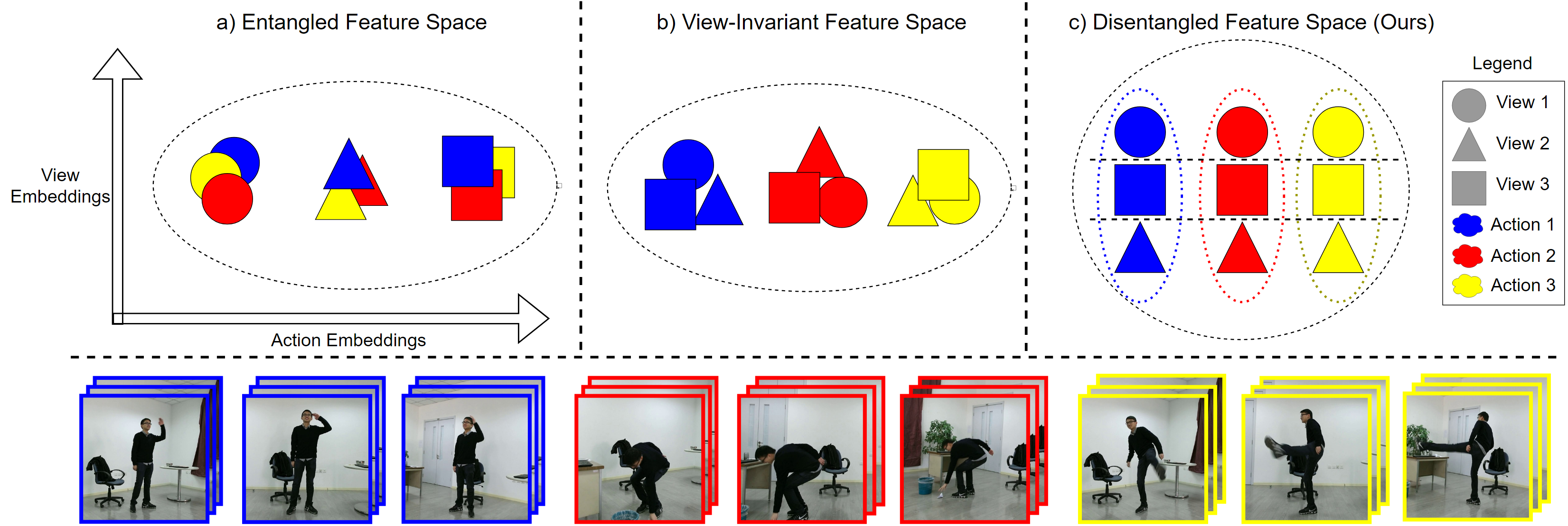}
    \caption{Conceptual visualization of learned action and view features plotted on the x- and y-axes, respectively, for different learning objectives. \textbf{a)} Traditional feature learning for multi-view action recognition may lead to view features becoming entangled in the action embedding space, causing different action classes to erroneously cluster due to similar viewpoints. \textbf{b)} View-invariant action features will properly cluster by action class, but the disentangled view features do not cluster properly. \textbf{c)} Our method disentangles the view features from the action features while still retaining the structure of both embedding spaces, improving performance on unseen viewpoints (see Fig. \ref{fig:tsne}).}
    \label{fig:motivation}

\end{figure*}

In specific regards to multi-view action recognition, it is currently a vital and heavily researched variant of HAR as it is more challenging and representative of real-world scenarios. For example, HAR datasets are commonly collected in an in-the-wild manner, where labelled actions are collected from viewpoints that are not explicitly controlled. %
Since single-view action recognition is usually the main learning objective in standard HAR, it is not guaranteed that the learned action representations of such models are robust to changes in viewpoints. 

However, in the previously mentioned real-world scenarios, videos of an action being performed are synchronously captured from multiple known viewpoints. Thus, multi-view action recognition has emerged as a more modern, challenging, and realistic version of HAR \cite{liu2016benchmarking,gao2019adaptive,cheng2012human,cai2014multi}. Additionally, many of these in-the-wild scenarios may lack the equipment or controlled collection that is required to capture additional modalities besides RGB frames, such as skeletal or depth \cite{sun2022human}. This proves to be a significant obstacle for current state-of-the-art multi-modal methods, as many of these solutions are heavily reliant on these modalities.

There are also many tangential areas of deep learning that affect the current approaches of multi-view action recognition. For example, a major obstacle in supervised deep learning is ensuring that learned features do not contain any irrelevant or confounding information with respect to the learning objective. This can be difficult to achieve since we cannot fully control how supervised models learn and represent information in their feature space. This can lead to learned features becoming 'entangled' with irrelevant information, thereby degrading performance. An example of this phenomenon in multi-view action recognition can be seen in Figure \ref{fig:motivation}, where view and action information can potentially become entangled in the feature space. However, large strides have been made in representation learning to close these gaps, including disentangled representation learning (DRL) \cite{liu2018detach,higgins2018towards}. DRL aims to improve the learned features for a training objective by enforcing strict separation between relevant and irrelevant information, improving overall performance.

Therefore, this work proposes a novel approach specifically towards multi-view action recognition in which we train a transformer decoder-based model for view-invariant action recognition using only a single modality. Our training method leverages disentangled representation learning by guiding our model to separate the action-relevant features from the view-relevant features. This enables our model to learn better quality action features that are robust to changes in viewpoints. We implement this disentanglement in two stages. In the first stage, we introduce the use of a singular view query in our transformer decoder to capture and disentangle view information from the learned action features. In the second stage, we instill relative comparisons between features using two supervised contrastive losses to assist in learning disentangled features. Standard cross-entropy learning only challenges a model to produce features that map a singular sample to the correct label. The model may implicitly learn the relative inter- and intra-class similarities and differences in the dataset, however this is not explicitly enforced and thus not guaranteed. On the other hand, supervised contrastive learning utilizes additional samples to guide a model to learn relatively identical features for similar samples (positives) and relatively disjoint features for dissimilar samples (negatives).  We show that our novel method of training, paired with our novel architecture, mitigates the degradation in multi-view action recognition performance stemming from the noise imparted by multiple viewpoints. Our main contributions are as follows:

\begin{itemize}
    \item A novel configuration of transformer decoder queries that enforce disentanglement of learned action and view features for view-invariant action recognition. 
    \item Two supervised contrastive losses and a query orthogonality loss to further supplement our disentangled representation learning.
    \item Unlike most methods, our approach is compatible with both the RGB or skeleton modality and is shown to outperform all unimodal state-of-the-art models on four different multi-view action recognition datasets.
\end{itemize}

\section{Related Work}
\label{sec:related_work}
\subsection{Multi-View Action Recognition}

Many recent action recognition works focus on tackling the problem of multi-view action recognition due to the modern and realistic challenge it poses in conjunction with its relevance in representation learning \cite{kong2017deeply,bian2023global}. The current dominant approaches consist of skeletal, RGB, or multi-modal methods; however, they are not without their limitations. Skeletal-based methods \cite{chen2021channel,shi2021adasgn,song2017end} are the most common approach to multi-view action recognition due to their quality representation of motion, removal of irrelevant information like background and clothing, and widespread availability of accurate skeleton ground truth labels in large-scale datasets \cite{liu2017pku,xia2012view,liu2019ntu}. Additionally, the compatibility of graph convolutional methods with skeleton-based action recognition led to widespread use in the literature \cite{chi2022infogcn,song2022constructing,cheng2020skeleton}. RGB-based action recognition is more sparse in the literature due to its lack of 3D structure, usually only being used in addition to other modalities \cite{wang2019generative,cheng2022spatial,das2020vpn}.
Contrary to these previous works, we demonstrate how our RGB-based model is able to achieve state-of-the-art multi-view action recognition performance, even over skeletal-based models. 

In addition to uni-modal approaches, multi-modal learning has shown promising results due to the additional quantity and modes of information provided during training. \cite{bruce2022mmnet} proposed a multi-modal approach to action recognition, named MMNet, in which they fuse skeletal and RGB-based features in a complementary manner to learn better action representations. 
MMNet does show impressive results on many multi-modal action recognition datasets, however we exhibit that we are able to achieve competitive results with only the RGB modality. Furthermore, our approach does not require the additional extraction of skeletal features, allowing for a more lightweight approach without any significant sacrifice in performance; an attribute not commonly seen with other RGB-based methods \cite{wang2018temporal,lin2020tsm,shah2023multi}. 
Previous RGB methods have similarly explored contrastive learning for multi-view action recognition, but still underperform when compared to skeletal-based models, whereas
our approach outperforms all uni-modal methods and sets multiple new state of arts. 

Learning view-invariant representations for multi-view action recognition has also been explored before for both uni- and multi-modal approaches \cite{li2018unsupervised,das2023viewclr,ji2021view,bian2023global}, and by using either solely convolutional or transformer-based architectures \cite{cheng2022spatial,vyas2020multi,ji2021view}. This segregated approach to architecture design may restrict the true potential of these models because of the respective limitations of CNNs and transformers, such as the quality of learned representations or speed of convergence. Our approach uses a hybrid architecture to limit each of these respective shortcomings, while also novelly introducing a specific configuration of queries for transformer decoders that facilitates feature disentanglement. We show that these novelties significantly contribute to our state-of-the-art results and stand as a strong improvement over previous works.

\subsection{Disentangled Representation Learning}

 Disentanglement has been applied in multi-view action recognition at both the representation learning level \cite{zhao2021learning,gao2022global,guo2022contrastive} and higher \cite{liu2020disentangling}, as well as other domains \cite{zhang2019gait,jin2022cloth,zhou2022human,zhou2022decoupling}. Similar to our method, \cite{tran2017disentangled} leveraged DRL for pose-invariance, but for face recognition. They also approach DRL from a joint generative-discriminative approach, which has been seen previously \cite{zheng2019joint}, in addition to purely generative methods \cite{karras2019style}. While our approach is purely discriminative, DRL's vast applicability across many domains is a testament to its capability to improve learned representations for specific tasks. However, improving learned representations is not limited to DRL. \cite{guo2022contrastive} explored improving skeleton-based action recognition representations using a self-supervised contrastive learning method. \cite{wang2018dividing}, somewhat similar to our method, uses view classification as an auxiliary task to assist in learning view-invariant representations.  

\section{Methodology}

\label{sec:approach}
\subsection{Motivation}

It is clear that the strongest factor of difficulty in multi-view action recognition, compared to general HAR, is the multiple viewpoints from which an action is captured. Supervised deep learning models trained on single-view action recognition datasets are not incentivized to learn as robust action representations as in the multi-view setting, since the performed actions are viewed from a static and homogeneous viewpoint. With the introduction of multiple viewpoints, the visibility and information of a performed action is variable, and thus requires more generalized and robust features to account for these perturbations. The relative ease at which humans are able to naturally perform multi-view action recognition has influenced the structure of all types of video - such as surveillance, sports, and entertainment - to be captured simultaneously from multiple viewpoints. In order to further advance general video understanding, it is imperative to develop supervised techniques that can process information from multiple viewpoints since even state-of-the-art single-view models will still fail in this facet.

\subsection{Our Approach}

\begin{figure*}
    \centering
    \includegraphics[width=\linewidth, height=9.5cm]{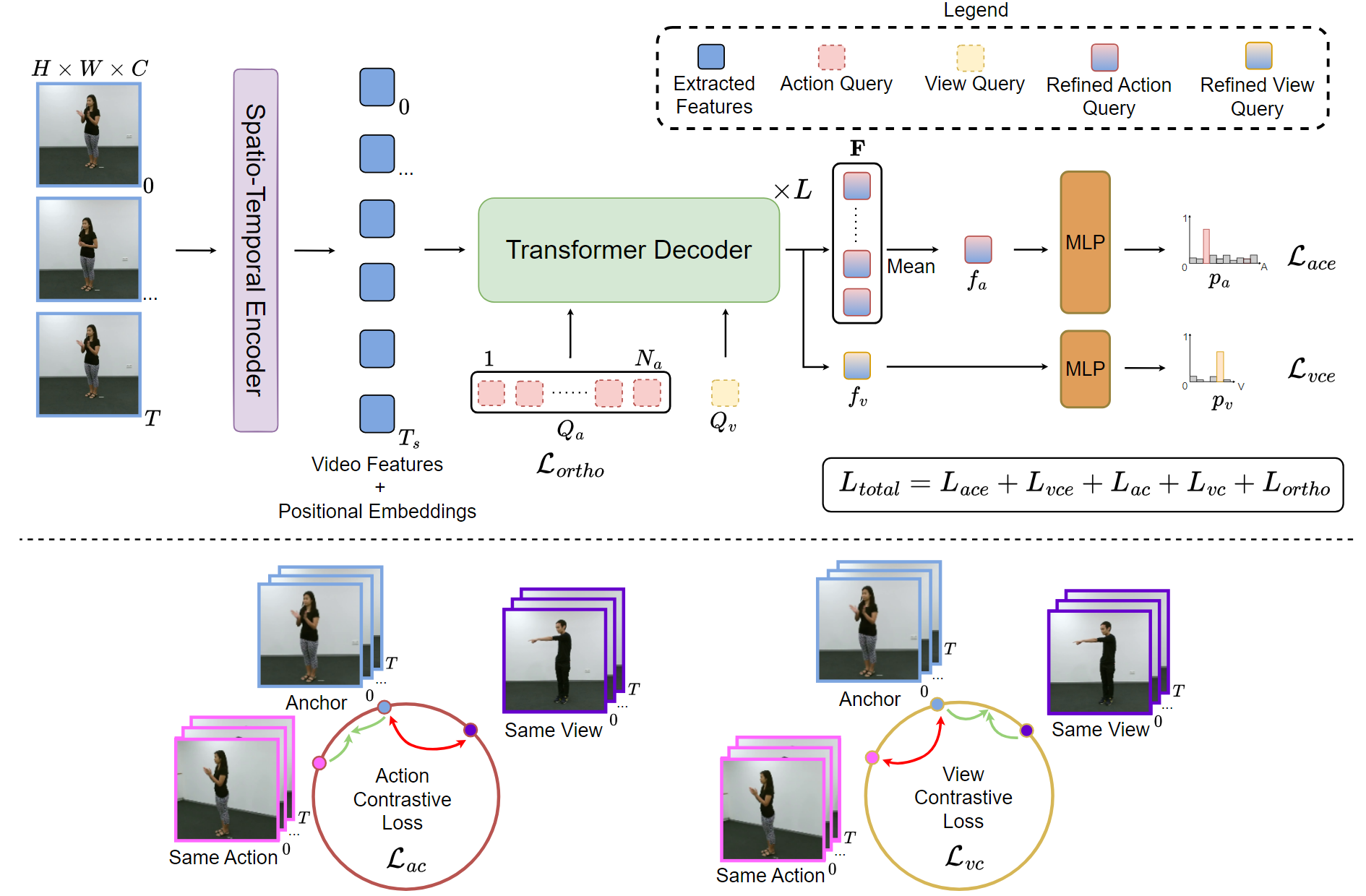}
    \caption{First, our spatio-temporal encoder extracts global features from a video. A transformer decoder is used to separately extract action and view features with the help of learnable action ($\bold{Q}_a$) and view ($\bold{Q}_v$) queries. In addition to the classification loss to predict the action and view, we also introduce contrastive loss to learn disentangled view and action representations.}
    \label{fig:model}

\end{figure*}

To this end, we propose DVANet to overcome these challenges, the full details of which can be seen in Figure \ref{fig:model}. We use a hybrid transformer decoder architecture with a 3D-CNN encoder to instill both local and global reasoning into our model. For the first stage of disentanglement, our decoder consists of multiple action queries to capture all action-relevant information, and a singular view query to capture all view-relevant information. In addition to standard cross-entropy loss, we utilize two supervised contrastive losses as a second stage of disentanglement to assist in the learning of view-invariant action features. These losses help drive feature disentanglement by enforcing the learned action and view features to only contain action and view information, respectively, and minimize leakage. We expect the action queries in the decoder to encapsulate all action-relevant information in a video without any perturbations due to changes in the camera view. Moreover, many previous methods rely on skeletal data in addition to RGB data to support the learning of view-invariant action features, whereas we achieve improved performance on the same task using only RGB data. We show that our novel architecture paired with our simple training method leads to significantly improved results over both RGB-based models and skeleton-based models alike on four different multi-view action recognition datasets: NTU RGB+D, NTU RGB+D 120, PKU-MMD, and N-UCLA. Specific model implementation details and all hyper-paramter values discussed in this section can be found in the supplement.

\paragraph{Input Videos}
During training, we pass a triplet of videos through our network. There is an anchor video, $\textbf{a}$, consisting of an action being performed from some known labeled viewpoint. We then retrieve a video of a different action captured from the same viewpoint as \textbf{a} and denote this video as $\textbf{sv}$ (same view). Lastly, we retrieve a video of the same action as \textbf{a} captured from a different viewpoint, denoted as \textbf{sa} (same action). We construct the triplets in this manner to provide positive and negative samples for our supervised contrastive feature learning, discussed later in this section. Providing additional positive and negative samples enables the model to gain a more holistic understanding of the learning objective through relative comparisons of samples, as opposed to standard single-label training.

\paragraph{Feature Extraction}
Our model consists of a backbone encoder paired with a transformer decoder. The backbone can be convolutional- or transformer-based, but an unfrozen, pretrained R3D backbone yielded best results. We firstly extract features using our backbone, which takes as input a video clip $\textbf{v} \in \mathbb{R}^{(T \times C \times H \times W)}$, where $T$ is the number of frames, and $C, H, $ and $W$ are the number of channels, height, and width of each frame respectively. The backbone encoder produces a feature map $\textbf{f} = ENC(\textbf{v}) \in \mathbb{R}^{(T_s \times D)}$ as output, where $D$ is the size of the hidden dimension and $T_s$ is the number of features for the entire video clip after temporal down-sampling. A learnable positional encoding is appended to this output, as is required, before it is passed to the transformer decoder for feature disentanglement.

\paragraph{Feature Disentanglement}
Our decoder takes the extracted spatio-temporal features from the backbone encoder as input and performs the first stage of disentanglement. The decoder consists of $N_a$ learnable action queries and a singular view query. Through our loss formulation, the action queries $\bold{Q}_a \in \mathbb{R}^{(N_a \times D)}$ learn to encapsulate the semantic information in a video that is pertinent only to the action being performed, irrespective of the view it was captured from. Contrastively, the singular view query $\bold{Q}_v \in \mathbb{R}^{(1 \times D)}$ is expected to only learn view-relevant information in a video, irrespective of the action being performed. This view query is meant to capture any perturbations that would otherwise occur in the action embedding space due to changes in the camera viewpoint, thereby 'disentangling' this noise from the learned action features. In addition to the contrastive learning, we use view classification as a training objective for this view query to learn view-relevant information. Thus, we are encouraging the action queries to produce view-invariant action features, since the view query captures any view-relevant information. 

The extracted features from our backbone encoder are then used as input for the transformer decoder, where these learnable queries are utilized to produce the decoder's output of refined features: $$\bold{F} = [\bold{f}_1, \cdots, \bold{f}_{N_a}] \in \mathbb{R}^{(N_a \times D)}, ~ \bold{f}_v \in \mathbb{R}^{(1 \times D)}$$  where $[\bold{f}_1, \cdots, \bold{f}_{N_a}]$ are the refined action features and $\bold{f}_{v}$ is the final learned view feature. We then take the average across the refined action features $\bold{F}$ to impart all learned query information into a final action feature, $\bold{f}_a$, where $\bold{f}_a \in \mathbb{R}^{(1 \times D)}$. Lastly, $\bold{f}_a, \bold{f}_v$, are passed to separate fully-connected linear layers $L_a, L_v$
that map the action and view features to their corresponding prediction dimensions, respectively. Action and view logits, $\bold{p}_a, \bold{p}_v$ are attained by $$\bold{p}_a = L_a(\bold{f_a}) \in \mathbb{R}^A \textup{ and } \bold{p}_v = L_v(\bold{f_v}) \in \mathbb{R}^V,$$ where $A$ and $V$ are the number of actions and views. 

Both the action and view branches of our model are tuned during training, but only the action branch is used during inference to produce action predictions. However, we cannot ensure that this architecture alone will result in disentangled action and view features without proper supervision. Therefore, we propose a combination of losses to properly enforce feature disentanglement. 

\paragraph{Loss Formulation}
 We incorporate standard cross entropy loss to train the features and linear layers using the corresponding action ($\mathcal{L}_{ace}$) or view ($\mathcal{L}_{vce}$) labels:
\begin{align}
  \mathcal{L}_{ace} = \sum_{n=1}^{N}\sum_{m=1}^M \text{log} \: \mathcal{P}_m(\bold{p}_a),
\end{align}

\begin{align}
  \mathcal{L}_{vce} = \sum_{n=1}^{N}\sum_{m=1}^M \text{log} \: \mathcal{P}_m(\bold{p}_v),
\end{align}
where $N$ is the number of training samples, $M$ is the number of actions (or the number of views in $L_v$), and $\mathcal{P}_m(\bold{p_*})$ is the probability of each predicted class from the logits with ground truth class $m$.

\paragraph{Contrastive Feature Loss}
Using the aforementioned triplet construction, we propose two opposing supervised contrastive learning objectives. Both of these losses use the same triplet construction, however the difference resides in the designation of positive and negative samples. 

Firstly, the action contrastive loss (Eq. \ref{actioncontrastive}) prioritizes learning view-invariant action features by treating the $\textbf{sa}$ video as a positive sample, and the $\textbf{sv}$ video as the negative sample. Thus, the distance between outputted features from the anchor video $\textbf{a}$, and a video of the same action from a different view $\textbf{sa}$, are minimized. Conversely, the distance between outputted features from $\textbf{a}$ and a video of a different action from the same viewpoint \textbf{sv} is maximized:
\begin{align}
    \label{actioncontrastive}
    \mathcal{L}_{ac} = \sum_{n=1}^N \: [\delta + \mathcal{D}(\bold{f}_a, \bold{f}_{sa}) - \mathcal{D}(\bold{f}_a, \bold{f}_{sv})],
\end{align}
where $\bold{f}_a, \bold{f}_{sa}, $ and $\bold{f}_{sv}$ denote the action feature outputs from the $\textbf{a}$, $\textbf{sa}$, and $\textbf{sv}$ videos, respectively, $\mathcal{D}(\cdot, \cdot)$ is a distance function, and $\delta$ is a margin parameter. The learned action features are then expected to become view-invariant, since they are encouraged to be identical when resulting from two different videos that consist of the same action, regardless of which viewpoint it was recorded from. We reverse the positive and negative samples for the view contrastive loss:

\begin{align}
    \label{viewcontrastive}
    \mathcal{L}_{vc} = \sum_{n=1}^N \: [\delta + \mathcal{D}(\bold{f}_a, \bold{f}_{sv}) - \mathcal{D}(\bold{f}_a, \bold{f}_{sa})],
\end{align}
 such that the singular view query in the decoder learns to produce similar features for videos that are from the same viewpoint, regardless of the action being performed. Since our main objective is to improve multi-view action recognition, training our model to disentangle the action-relevant features from the view-relevant features in this manner provides cleaner, more accurate action representations that are robust to changes in the camera viewpoint.

\paragraph{Orthogonal Query Loss}
As seen in previous works \cite{fei2022vit,zhang2021orthogonality}, initializing and enforcing orthogonality between queries in the decoder improves  both model performance and numerical stability. Leveraging orthogonal queries aligns with our learning objective as we desire each action query to learn distinct, non-overlapping semantic information in the video to avoid redundancy. We use the cosine similarity between action queries $(\bold{q}_m, \bold{q}_n \in Q_a)$ to compute the loss and discourage any overlap between action queries during training:

\begin{align}
    \label{ortholoss}
    \mathcal{L}_{ortho} = \sum_{m=0}^{N_a}\sum_{n \neq m} \: \left|\frac{\langle \bold{q}_m, \bold{q}_n\rangle} {||\bold{q}_m||\:  ||\bold{q}_n||}\right|.
\end{align}

Our overall loss is then calculated as follows:
\begin{align*}
    \mathcal{L} = \mathcal{L}_{ace} +  \mathcal{L}_{vce} + \mathcal{L}_{ac} +  \mathcal{L}_{vc} +  \mathcal{L}_{ortho}.
\end{align*}

\section{Experimental Evaluation}

\subsection{Results}
We show results on three large-scale, multi-view action recognition datasets (NTU RGB+D, NTU RGB+D 120, PKU-MMD) and one smaller scale dataset (N-UCLA) to exhibit the effectiveness of our approach with varying amounts of data. All of these datasets provide skeletal and depth information in addition to the RGB frames, however we only utilize RGB information in our main approach. We show results on the standard cross-subject and cross-view evaluation protocols provided for each dataset. Additional analyses and details regarding results, using multiple modalities, and dataset information are in the supplement. 

\paragraph{NTU RGB+D \& NTU RGB+D 120}
In Table \ref{tab:NTU60results}, we provide a comparison of action test accuracy on NTU RGB+D with previous state-of-the-art (SOTA) unimodal models. In both the cross-subject and cross-view evaluation protocol, we achieve the highest performance and beat the previous best (which used the skeleton modality) in each protocol by 0.2\% and 0.7\%, respectively. When only compared to RGB models, we beat the previous best by an impressive 1.5\% margin in cross-subject, and 0.2\% in cross-view. 
An important distinction of our method is the use of view classification to better learn and identify view-relevant information, thereby making it easier to disentangle from our model's learned action features (see Table \ref{tab:lossablation}).

In Table \ref{tab:NTU120results}, we observe similar superior performance of our method when applied to the larger dataset of NTU RGB+D 120. 
Our model achieves SOTA results compared to other unimodal models, achieving a 0.5\% and 0.4\% improvement over the SOTA skeleton-based model in the cross-subject and cross-setup protocols, respectively. More importantly, we significantly outperform RGB based models with large improvements of 4.8\% and 4.1\%, respectively. Smaller improvements are seen over skeleton-based methods due to the difference in quality of information between modalities, whereas our approach yields significantly large improvements over other RGB methods.

\paragraph{PKU-MMD}
Table \ref{tab:PKresults} shows the comparisons of our results with previous works on PKU-MMD. Again we report large margins between our model and previous SOTA methods, with improvements of 2.2\% and 1\% on cross-subject and cross-view, respectively. The listed RGB-based methods even use additional depth information and still do not reach the our level of performance. Our model also beats all baselines in Table \ref{tab:PKresults} when we train using the skeleton modality, which can be found in the supplement.

\paragraph{N-UCLA}
Compared to the other three large-scale datasets, N-UCLA is considerably smaller in both the number of videos and actions. Some works \cite{shah2023multi} evaluate their model using transfer learning and fine-tuning on this dataset, while others train fully from scratch.  Previous works such as \cite{vyas2020multi} are only able to obtain competitive results when using transfer learning, as training from scratch resulted in severely degraded performance. We train our model from scratch on this dataset and beat recent RGB works by a significant margin of 2.3\% and 4.8\%, as seen in Table \ref{tab:N-UCLAresults}.  Our results indicate that our approach does not require mass amounts of data to achieve SOTA performance.

\begin{table}[]
\centering
\resizebox{\linewidth}{!}{
\begin{tabular}{l|l|l|l}
\Xhline{2\arrayrulewidth}
\multicolumn{1}{c|}{\textbf{Model}} & \multicolumn{1}{c|}{\textbf{Modality}} & \multicolumn{1}{c|}{\textbf{Cross-Subject}} & \multicolumn{1}{c}{\textbf{Cross-View}} \\
\Xhline{2\arrayrulewidth}
\cite{liu2020disentangling} (CPVR '20) & Skeleton & 91.5 & 96.2 \\ 
\cite{chen2021channel} (ICCV '21) & Skeleton & 92.4 & 96.8 \\
\cite{chi2022infogcn} (CVPR '22) & Skeleton & 93.0 & 97.1\\
\cite{trivedi2023psumnet} (ECCV '22 W) & Skeleton & 92.9 & 96.7\\
\cite{duan2022pyskl} (ACM MM '22) & Skeleton & 92.6 & 97.4\\
DG-STGCN (ACM MM '22) & Skeleton & 93.2 & 97.5\\
EfficientGCN (PAMI '22) & Skeleton & 92.1 & 96.1\\
\cite{zhang2023hierarchical} (AAAI '23) & Skeleton & 90.4 & 95.7 \\
\Xhline{2\arrayrulewidth}
\cite{wang2018dividing} (ECCV '18) & RGB & 63.3 & 70.6 \\
\cite{baradel2018glimpse} (CVPR '18) & RGB & 86.6 & 93.2\\
\cite{vyas2020multi} (ECCV '20) & RGB & 82.3 & 86.3 \\
\cite{piergiovanni2021recognizing} (CVPR '21) & RGB & - & 93.7\\
\cite{cheng2022spatial} & RGB & 91.9 & 95.4\\
\cite{das2023viewclr} (WACV '23) & RGB & 89.7 & 94.1\\
\cite{shah2023multi} (WACV '23) & RGB & 91.4 & \textit{98.0}\\
\Xhline{3\arrayrulewidth}
\textbf{DVANet} & RGB &  {\textbf{93.4}} \underline{(+1.5)} & {\textbf{98.2}} \underline{(+0.2)} \\

\end{tabular}
}
\caption{Comparison of our model with previous SOTA unimodal models on NTU RGB+D. Highest accuraices are in bold, and margin of improvement over RGB methods are underlined.}
\label{tab:NTU60results}

\end{table}

\begin{table}[]
\centering
\resizebox{\linewidth}{!}{
\begin{tabular}{l|l|l|l}
\Xhline{2\arrayrulewidth}
\multicolumn{1}{c|}{\textbf{Model}} & \multicolumn{1}{c|}{\textbf{Modality}} & \multicolumn{1}{c|}{\textbf{Cross-Subject}} & \multicolumn{1}{c}{\textbf{Cross-Setup}} \\
\hline

 \cite{liu2020disentangling} (CPVR '20) & Skeleton & 86.9 & 88.4 \\ 
 \cite{chen2021channel} (ICCV '21) & Skeleton & 88.9 & 90.6 \\ 
\cite{chi2022infogcn}  (CVPR '22) & Skeleton & {89.8} & {91.2}\\
\cite{chi2022infogcn}  (CVPR '22) & Skeleton & 88.5 & 89.7\\
 \cite{song2022constructing} (PAMI '22) & Skeleton & 88.7 & 88.9\\
 \cite{duan2022pyskl} (ACM MM '22) & Skeleton & 88.6 & 90.8\\
\cite{trivedi2023psumnet} (ECCV '22 W) & Skeleton & 89.4 & 90.6\\
\cite{zhang2023hierarchical} (AAAI '23) & Skeleton & 85.6 & 87.5 \\
\Xhline{2\arrayrulewidth}

 \cite{das2023viewclr} (WACV '23) & RGB & 84.5 & 86.2\\
 \cite{shah2023multi} (WACV '23) & RGB & 85.6 & 87.5\\
\Xhline{3\arrayrulewidth}

\textbf{DVANet} & RGB & {\textbf{90.4}} \underline{(+4.8)} & {\textbf{91.6}} \underline{(+4.1)} \\
\end{tabular}
}
\caption{Comparison of our model with previous SOTA unimodal models on NTU RGB+D 120. Highest accuraices are in bold, and margin of improvement over RGB methods are underlined.}
\label{tab:NTU120results}
\end{table}

\begin{table}[]
\centering
\large
\resizebox{\linewidth}{!}{
\begin{tabular}{l|l|l|l}
\Xhline{2\arrayrulewidth}
\multicolumn{1}{c|}{\textbf{Model}} & \multicolumn{1}{c|}{\textbf{Modality}} & \multicolumn{1}{c|}{\textbf{Cross-Subject}} & \multicolumn{1}{c}{\textbf{Cross-View}} \\
\hline
\cite{song2017end}  & Skeleton & 86.9 & 92.6\\
\cite{li2018co} (IJCAI '18) & Skeleton & 92.6 & {94.2}\\
\cite{elias2019understanding} (ISM '19) & Skeleton & 86.5 & 92.2 \\

\Xhline{2\arrayrulewidth}
\cite{wang2018temporal} (PAMI '19) & RGB+Depth & 85.0 & 85.7\\
\cite{lin2020tsm} (PAMI '22) & RGB+Depth & 91.7 & 92.6\\
\cite{cheng2022spatial} (TSN) & RGB+Depth & 92.6 & 92.1\\
\cite{cheng2022spatial} (TSM) & RGB+Depth & {93.6} & {94.2}\\

\Xhline{3\arrayrulewidth}
\textbf{DVANet} & RGB & {\textbf{95.8}} \underline{(+2.2)} & {\textbf{95.2}} \underline{(+1.0)} \\
\end{tabular}
}

\caption{Comparison of our model with previous SOTA models on PKU-MMD.  Highest accuraices are in bold, and margin of improvement over RGB methods are underlined.}
\label{tab:PKresults}
\end{table}

\begin{table}[]
\centering
\resizebox{\linewidth}{!}{
\begin{tabular}{l|l|l|l}
\Xhline{1\arrayrulewidth}
\multicolumn{1}{c|}{\textbf{Model}} & \multicolumn{1}{c|}{\textbf{Modality}} & \multicolumn{1}{c|}{\textbf{Cross-Subject}} & \multicolumn{1}{c}{\textbf{Cross-View}} \\
\hline

\cite{wang2018dividing} (ECCV '18) & RGB & {92.1} & 86.5 \\
\cite{baradel2018glimpse} (CVPR '18) & RGB & - & 87.6\\
\cite{vyas2020multi} (ECCV '20) & RGB & 87.5 & 83.1 \\
\cite{das2023viewclr} (WACV '23)  & RGB & - & 89.1\\
\cite{shah2023multi} (WACV '23)  & RGB & - & {91.7}\\

\Xhline{3\arrayrulewidth}
\textbf{DVANet} & RGB & {\textbf{94.4}} \underline{(+2.3)} & {\textbf{96.5}}  \underline{(+4.8)} \\
\end{tabular}
}

\caption{Comparison of our model with other unimodal models on N-UCLA.  Highest accuraices are in bold, and margin of improvement over RGB methods are underlined.}

\label{tab:N-UCLAresults}
\end{table}

\section{Discussion \& Analysis}

To provide deeper insights into our approach, we provide multiple ablations and analyses on our losses and results. All ablations are performed using the cross-subject evaluation protocol on PKU-MMD, since PKU-MMD is of moderate size. We use the cross-subject protocol since our model is trained in a supervised manner for view-invariant action recognition from seen viewpoints. However, Fig. \ref{fig:view-tsne} does show that our model implicitly learns features that can perform on unseen viewpoints. Additional details regarding protocols, Fig. \ref{fig:view-tsne}, and performance on unseen viewpoints can be found in the supplement.

\subsection{Ablations}
To measure each of our losses' contributions to the learning objective, we ablate each loss during training in Table \ref{tab:lossablation}.

\paragraph{Effectiveness of Using View Information}
When we remove the cross-entropy and contrastive losses related to view information, we observe a significant 1.8\% drop in performance in the first row of Table \ref{tab:lossablation}. These two losses aid in the disentanglement of view information from the action features, and thus removing them expectedly degrades the quality of learned action features. With most state-of-the-art methods already achieving extremely high results on PKU-MMD, a nearly 2\% drop in performance does have a notable affect on the margin with which we improve over previous methods. This shows that multiple viewpoints does in fact perturb learned action features enough to drop performance.

\paragraph{Effectiveness of Contrastive Losses}
Rows 2-4 of Table \ref{tab:lossablation} show the individual and combined contributions of our two contrastive losses. We see around a 0.5\% decrease in performance when only ablating one contrastive loss, which may be due to the other contrastive loss still assisting the model in improving its learned features through relative comparisons. However, performance drops by nearly 2\% when removing both contrastive losses, exhibiting that constructing triplets for contrastive learning during training is superior to simple single-label cross-entropy learning for this task.

\paragraph{Effectiveness of Orthogonal Loss}
As described in the methodology, orthogonal queries are forced to learn non-overlapping features to reduce redundancy and provide additional numerical stability. The last row in Table \ref{tab:lossablation} shows that we obtain a roughly 1\% increase when using the orthogonal loss. The largest drops in performance occur from removing either the contrastive losses or view information during training, indicating that these losses contribute more directly to our learning objective than orthogonal loss.

\begin{table}[]
\centering
\begin{tabular}{lllll|l}
\Xhline{2\arrayrulewidth}
\multicolumn{1}{c}{$\mathcal{L}_{ace}$} &\multicolumn{1}{c}{$\mathcal{L}_{vce}$} & \multicolumn{1}{c}{$\mathcal{L}_{ac}$} & \multicolumn{1}{c}{$\mathcal{L}_{vc}$} & \multicolumn{1}{c|}{$\mathcal{L}_{ortho}$} & \multicolumn{1}{c}{\textbf{Cross-Subject}} \\
\hline
{\cmark} & {\xmark} & {\cmark} & {\xmark} & {\cmark}  & 94.0\\
{\cmark} & {\cmark} & {\xmark} & {\cmark} & {\cmark}  & 95.3 \\
{\cmark} & {\cmark} & {\cmark} & {\xmark} & {\cmark}  & 95.2\\
{\cmark} & {\cmark} & {\xmark} & {\xmark} & {\cmark}  & 94.2 \\
{\cmark} & {\cmark} & {\cmark} & {\cmark} & {\xmark}  & 95.0\\

\Xhline{2\arrayrulewidth}
{\cmark} & {\cmark} & {\cmark} & {\cmark} & {\cmark} & \textbf{95.8}\\
\Xhline{2\arrayrulewidth}
\end{tabular}
\caption{Results on PKU-MMD when certain losses are ablated during training. When all losses are used, we achieve the highest performance.}
\label{tab:lossablation}

\end{table}

\subsection{Analysis}

\begin{figure*}[ht]
    \centering
    \includegraphics[height=6.5cm]{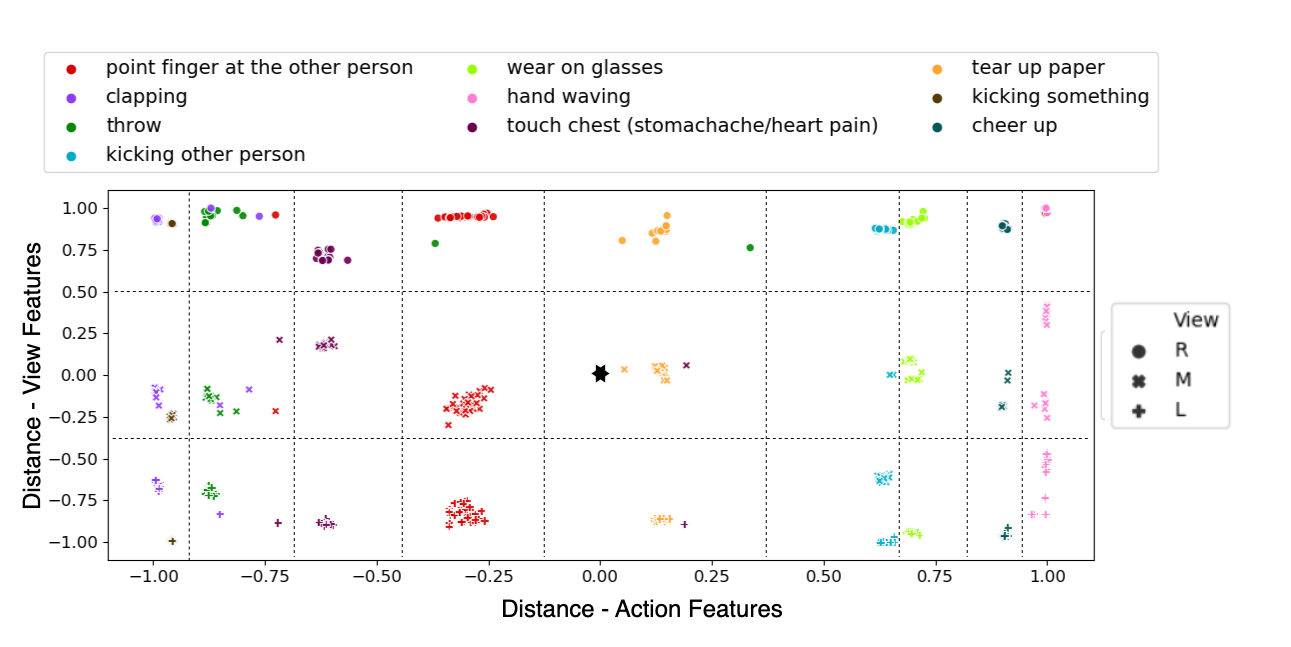}

    \caption{Visualizations of view and action features to show feature disentanglement. Each point is plotted by the distance of its respective action and view features from the reference point ($\ast$). The learned action features form clusters by action class, and within each cluster are separated by viewpoint. Grid lines are added to help show the segregation of clusters. 
    }

    \label{fig:tsne}
\end{figure*}

\paragraph{Visualizing Disentanglement}
To further support our claim of disentangled features, Figure \ref{fig:tsne} shows the clustering of the learned action and view features for the 10 most common actions in the PKU-MMD cross-subject test set. A single test sample is randomly selected as a reference point (shown by $\ast$). We then calculate the distance between its action feature and the action features of every other sample, and do the same for the view features. The x-axis denotes the distance of a given sample's action feature with respect to our reference point's action feature, with the y-axis denoting the same for the view features. We see that each action class (denoted by color) clusters separately along the x-axis from other action classes in the feature space. Additionally, for each action class there are three distinct clusters along the y-axis: one for each viewpoint. The deduction from these observations is two-fold: the instances of a certain action class cluster together along the x-axis, indicating that our model produces action features that are distinguishable from other action classes. Secondly, instances of the same action from different viewpoints are clustered separately along the y-axis. Thus, the learned view features of our model were successfully disentangled from the action features.

\begin{figure}[hbt!]
    \centering
    \includegraphics[height=4cm, width=\linewidth]{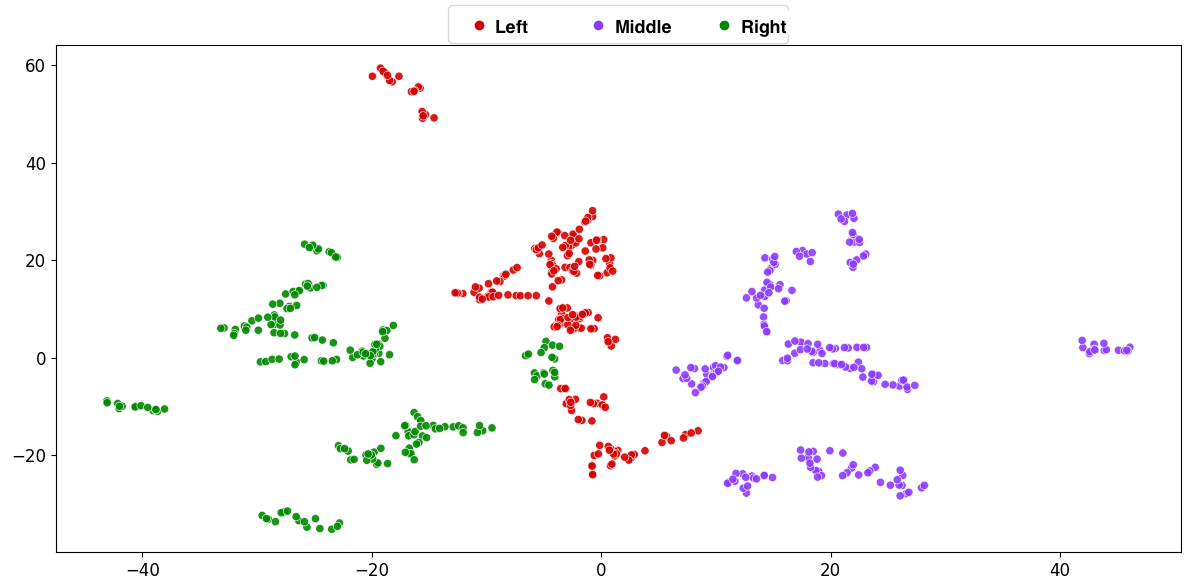}
    \caption{t-SNE of the view features from our cross-view model evaluated on the cross-subject test set. The 'left' viewpoint (shown in red) is unseen during training and is still separated from the known views during testing.}
    \label{fig:view-tsne}
\end{figure}

\paragraph{Supplement}
The supplement provides many additional details and discussions for analyses shown in this paper, such as Figure \ref{fig:view-tsne}. More results, comparisons, ablations, and visualizations are also provided, such as training our model with skeletal or multi-modal data.

\section{Conclusion}

We propose a novel transformer decoder-based architecture in tandem with two supervised contrastive losses for multi-view action recognition. By disentangling the view-relevant features from action-relevant features, we enable our model to learn action features that are robust to change in viewpoints. 
We show through various ablations, analyses, and visualizations that changes in viewpoint impart perturbations on learned action features. Thus, disentangling these perturbations improves overall action recognition performance. Uni-modal state-of-the-art performance is attained on four large-scale multi-view action recognition datasets, highlighting the efficacy of our method.

\section{Acknowledgements}
This research is based upon work supported in part by the Office of the Director of National Intelligence (ODNI), Intelligence Advanced Research Projects Activity (IARPA), via  2022-21102100004. The views and conclusions contained herein are those of the authors and should not be interpreted as necessarily representing the official policies, either expressed or implied, of ODNI, IARPA, or the U.S. Government. The U.S. Government is authorized to reproduce and distribute reprints for governmental purposes notwithstanding any copyright annotation therein.

\bibliography{aaai24}

\end{document}